# Dynamic Adjustment of the Motivation Degree in an Action Selection Mechanism

**Carlos Gershenson***, **, ***
**Fundación Arturo Rosenblueth
***Facultad de Filosofía y Letras/UNAM
carlos@jlagunez.iquimica.unam.mx

**Pedro Pablo González***
*Instituto de Química/UNAM
Ciudad Universitaria, 04510
México, D. F. México
ppgp@servidor.unam.mx

## Abstract

This paper presents a model for dynamic adjustment of the motivation degree, using a reinforcement learning approach, in an action selection mechanism previously developed by the authors. The learning takes place in the modification of a parameter of the model of combination of internal and external stimuli. Experiments that show the claimed properties are presented, using a VR simulation developed for such purposes. The importance of adaptation by learning in action selection is also discussed.

## 1. Introduction

In the context of the behaviour-based systems (Brooks, 1986; Brooks , 1989; Maes, 1990; Maes, 1994), the action selection problem refers to what action the agent (robot, animat or artificial creature) must select every moment in time; whereas an action selection mechanism (ASM) specifies how these actions are selected. An ASM is a computational mechanism that must produce as an output a selected action, when different external and/or internal stimuli have been given as inputs.

The great majority of the ASM proposed in the literature (Tyrrell, 1993), include an additive model for stimuli combination. The stimuli addition could be interpreted in many ways, but the essence is that this type of stimuli combination models a behaviour with a more reactive tendency than a motivated one. This can be explained in the following terms: for high values of the external inputs, the additive combination could surpass some value threshold previously established and, therefore, trigger a certain external action, still for non significant values of the internal inputs.

On the other hand, in a pure model of multiplicative combination both types of inputs are necessary so that an external consummatory action of the entity can be executed. This is, if we considered that ${}_jS_j$ is the sum of all the associated external inputs to the internal state $E_i$, then ${}_jS_j * E_i \neq 0$ only if ${}_jS_j \neq 0$ and $E_i \neq 0$. Pure multiplicative models have a limit ratio to very small values of the external inputs and/or of the internal inputs, since it does not matter how great is the external input if the internal input is near zero and vice versa.

An alternative to eliminate the undesired effects of the pure additive and multiplicative combinations is to consider a model that incorporates the most valuable elements of both types of combination. This is achieved by the combination model of an ASM developed by us (González, 2000).

Adaptation is one of the desirable characteristics in the action selection of an autonomous agent. The adaptation in an ASM can be obtained from three main approaches: preprogrammed adaptive behaviours, learned adaptive behaviours and evolved adaptive behaviours (Meyer and Guillot, 1990). Adaptation by learning is a process by which an autonomous agent is able to adjust its behaviour to different types of environmental changes. An autonomous agent is adaptive by learning if it has skills, which allow to improve its performance in time (McFarland, 1990).

In this paper we discuss how the adaptation by learning improves the action selection in an autonomous agent, from an ASM developed by the authors (González, 2000; González et. al., 2000; Gershenson et. al., 2000). Two types of adaptation by learning are present in this ASM: (1) classical conditioning, which allows that new behaviours and emergents properties to arise in the action selection, increasing its adaptive level, and (2) reinforcement learning approach (Humphrys, 1996), which allows the motivation degree in the action selection to be adjusted dynamically. This paper discusses the second type of adaptation by learning. This type of learning takes place over a parameter belonging to the model for combination of external and internal stimuli developed by us (González et. al., 2000) and used by the ASM. This parameter allows to regulate the dependence degree of the external behaviour executed by the agent from its internal states, and its current value is learned dynamically.

The paper is organized as follows: The next section briefly describes the ASM developed by us, the Behavioural Columns Architecture, doing special emphasis in the basic functional unit of the ASM, the internal behaviour. Section 3 presents the structure and functioning of the *intero/extero/drive congruence* internal behaviour, where the combination of external and internal signals is carried out. Section 4 discusses the model for combination of external and internal stimuli used by the ASM, and the role of the parameter that allows to regulate the motivation degree in this combination. In section 5 we present and discuss the learning process by which the motivation degree in the action selection can be adjusted dynamically. Finally, section 6 presents the simulation and experiments developed to verify the role of this type of adaption by learning in the action selection.

## 2. Behavioural Columns Architecture

The ASM developed by us has been structured from a network of blackboard nodes developed by the authors (González and Negrete, 1997; Negrete and González, 1998). A blackboard node is integrated by the following components: a set of independent modules called knowledge sources, which have specific knowledge about the problem domain; the blackboard, a shared data structure through which the knowledge sources communicate to each other by means of the creation of solution elements on the blackboard; the communication mechanisms, which establish the interface between the nodes and a node and the external and internal mediums; the activation state registers of the knowledge sources; and a control mechanism, which determines the order in that the knowledge sources will operate on the blackboard.

The developed action selection mechanism was named Behavioural Columns Architecture (BeCA), which is an extension of the Internal Behaviour Network (IBeNet) (González, 2000). As it can be appreciated in Figure 1, the actual architecture of the ASM exhibits two blackboard nodes: the cognitive node and the motivational node.

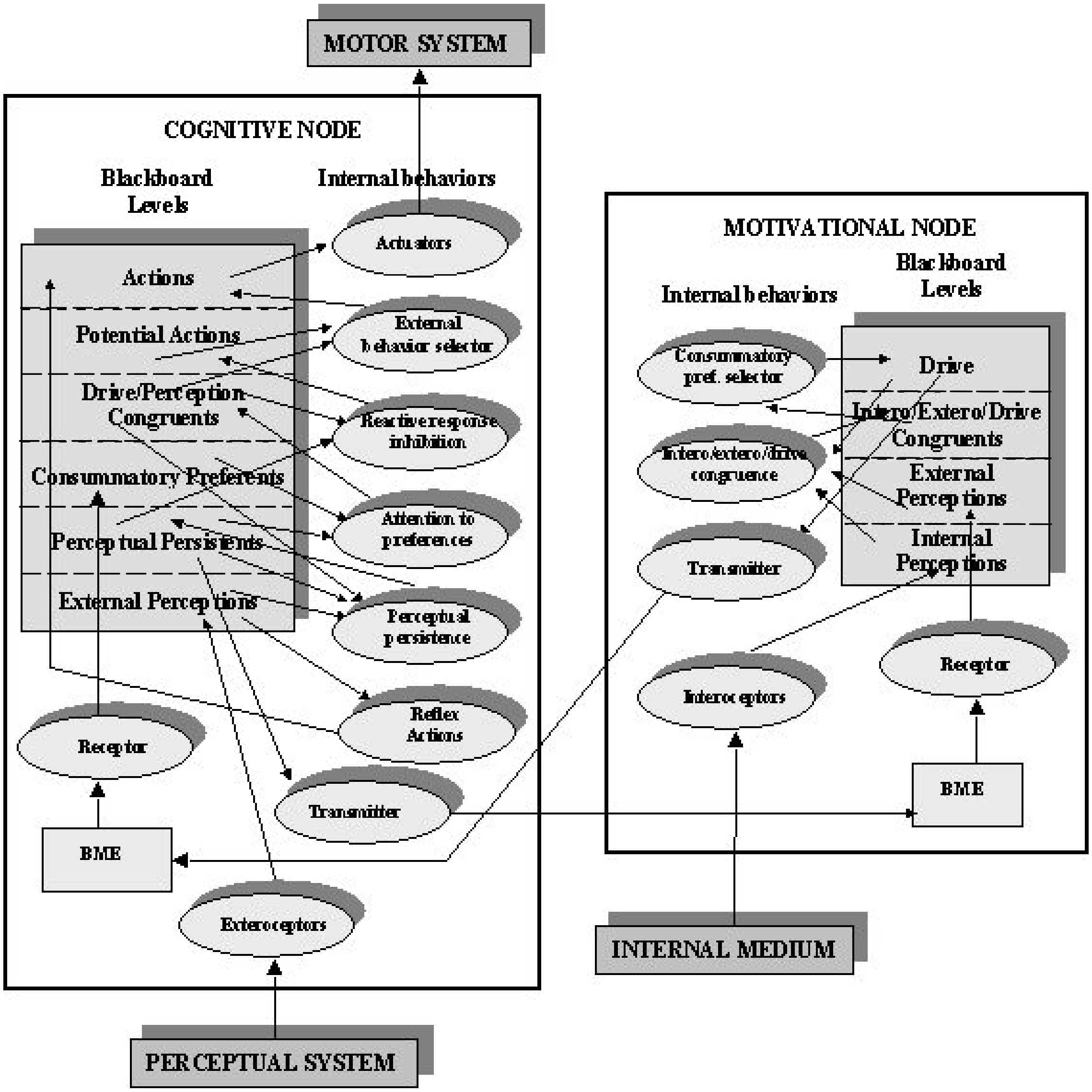

Figure 1. Behavioural Columns Architecture

The basic functional unit of the BeCA is the "internal behaviour". The term "internal behaviour" has been used to describe the processes of creation and modification of solution elements that occur on the blackboard of the node. Structurally, an internal behaviour is a package of elemental behaviours, also known as production rules (if <condition> then <action>). Elemental behaviours of different kinds are organized forming "behavioural columns" that cross vertically different blackboard levels.

The structure of an elemental behaviour has three basic elements: a parameters list, a condition part, and an action part. The parameters list specifies which are the condition elements, the action elements, and the coupling strengths related with the elemental behaviour. The condition of an elementary behaviour describes the configuration of solution elements on the blackboard that is necessary, so that the elementary behaviour contributes to the solution processes of the problem. The way in which an elemental behaviour contributes to the solution of the problem is specified in its action, which can consist of the creation or modification of solution elements in certain blackboard levels.

A coupling strength is represented by a vector Fa = ($Fa_{i1}$, $Fa_{i2}$, ..., $Fa_{in}$) of n real components, where each of these components represent the efficacy with which the elemental behaviour i can satisfy the condition j. Depending on the nature of the elemental behaviour, the components of the vector Fa may have fixed or modifiable values. The existence of modifiable coupling strengths allows the incorporation of learning processes to improve the action selection, making it more adaptive. The vector Fa of coupling strengths is to an elemental behaviour the same as a weight vector to an artificial neuron.

The structure of a blackboard node in BeCA consists of five basic elements: the internal behaviours, the blackboard, the activity state registers of the internal behaviours, the interface/communication mechanisms, and the competition mechanism.

The cognitive node blackboard is structured in six levels of abstraction: external perceptions, perceptual persistents, consummatory preferents, drive/perception congruents, potential actions, and actions. For the cognitive node, the following internal behaviours have been defined: *perceptual persistence, attention to preferences, reactive response inhibition,* and *external behaviours selector*. The defined mechanisms of communication for this node are the *exteroceptors* and *actuators*, which establish the interface between the cognitive node and external medium; and the *receptor* and *transmitter* mechanisms, which establish communication with the motivational node. The role of the cognitive node includes the processes of representation of the perceptual signals, integration of internal and external signals, inhibition of reactive responses, and the selection of the external behaviour that adjusts better to the actual external conditions and internal needs.

The blackboard of the motivational node is structured in four abstraction levels: internal perceptions, external perceptions, intero/extero/drive congruents, and drive. The internal behaviours that operate in the motivational node are the following: *intero/extero/drive congruence* and *consummatory preferences selector.* The communication of this node is established from the *interoceptors,* which define the interface between the node and the internal medium; and the *receptor* and *transmitter* mechanisms, which establish the communication with the cognitive node. The role of the motivational node includes the processes of representation of internal signals, combination of internal and external signals and the selection of the more appropriate consummatory preference. A full discussion of the model for combination of external and internal stimuli, used by the motivational node in BeCA, can be found in (González et. al., 2000).

## 3. The *intero/extero/drive congruence* internal behaviour

At the level of the *intero/extero/drive congruence* internal behaviour of the motivational node is carried out the combination of external and internal signals to determine the action that must be executed by the entity. These signals are registered in the external perceptions and internal perceptions levels, respectively, of the motivational node. The solution elements registered in the external perceptions level are originated by unconditional stimuli (US), conditional stimuli (CS), or both; all of them initially projected in the external perceptions level and later recovered from the perceptual persistents level of the cognitive node. On the other hand, the solution elements registered in the internal perceptions level of the motivational node represent the values of the internal states that are able to satisfy part of the conditions of certain *intero/extero/drive congruence* elementary behaviours that operate on this level.

Whenever the condition of an elemental behaviour is satisfied, this executes its final action, creating the solution element $C_i$ with certainty $A_i^C$ in the intero/extero/drive congruents level of the blackboard of the motivational node. The level of certainty $A_i^C$ represents the activity of the elemental behaviour i. Figure 2 shows the structure of an elemental behaviour of this type.

**Parameters:**
**condition solution elements: $E_i$, $S_j$, $D_i$ (j = 1, 2, ...)**
**action solution elements: $C_i$**
**coupling strengths: $Fa_{ij}^S$ (j = 1, 2, ...)**
**importance attributable to the internal state: $_i$**
**Condition:**
**( $_i$ = 0 AND**
**blackboard-level = Internal Perceptions AND**
**cert($E_i$, $O_i^E$) AND**
**$O_i^E$ 0 AND**
**blackboard-level = External Perceptions AND**
**cert($S_i$, $O_i^S$) AND**
**$O_i^S$ 0) OR**
**( 0 AND**
**blackboard-level = Internal Perceptions AND**
**cert($E_i$, $O_i^E$) AND**
**$O_i^E$ 0 AND**
**blackboard-level = External Perceptions AND**
**cert($S_i$, $O_i^S$))**
**Action:**
**calculate $A_i^C = O_i^E*( _i + {}_jFa_{ij}^S*O_j^S) + O_i^D$ AND**
**blackboard-level = Propio/Extero/Drive Congruents AND**
**create-solution-element ($C_i$, $A_i^C$)**

Figure 2. Structure of an *intero/extero/drive congruence* elementary behaviour

# 4. The model for combination of internal and external stimuli

The model for combination of internal and external stimuli used by *intero/extero/drive congruence* internal behaviour is the one given by expression (1).

$$A_i^C = Fa_i^E O_i^E\left(\alpha_i + \sum_j Fa_{ij}^S O_j^S\right) + Fa_i^D O_i^D \qquad (1)$$

where $A_i^C$ is the new activation level correspondent to the intero/extero/drive congruents level of the motivational node; $O_i^E$ is the signal from the internal perceptions level and $Fa_i^E$ its coupling strength; $O_j^S$ is the signal from the external perceptions level and $Fa_{ij}^S$ its coupling strength; $O_i^D$ is the signal from the drive level, and $Fa_i^D$ its coupling strength; and $\alpha_i$ regulates the combination of the internal and external signals.

## 4.1. Case 1: $\alpha_i = 0$

When $_i$ is equal to zero, the *intero/extero/drive congruence* elemental behaviour will be able to activate only when the internal and external signals ($O_i^{E1}$ 0 and ${}_jFa_{ij}^S*O_j^{S1}$ 0) associated to these coincide. If the necessary internal state is not present, then the external signal is not sufficient by itself to evoke the motor action, since the *attention to preferences* internal behaviour is activated only when their external signal (solution element in the perceptual persistent level of the cognitive node) and the drive signal (solution element in the consummatory preferents level of the cognitive node) coincides. On the other hand, although a very strong internal signal exists, corresponding to an urgent necessity of the entity, if the usable external signal is not present, the *intero/extero/drive congruence* elemental behaviour will not activate. Then, when $_i$ = 0, both factors (external signal and internal signal) are necessary so that the *intero/extero/drive congruence* internal behaviour will be activated. Physiologically this can be interpreted, when thinking that the inputs that come from the internal states sensitize this internal behaviour with the signals that come from the external medium.

## 4.1. Case 2: $\alpha_i \neq 0$

When $_i$ is different of zero, more weight (or importance) is being granted to the internal state than to the external inputs. So, still in total absence of external inputs, the *intero/extero/drive congruence* elemental behaviour could be activated for a very strong value of the internal signals (when the value of $_i$ is near one). Thus, when the internal need is extremely high and external signals do not exist, the *intero/extero/drive congruence* elemental behaviour can be activated, and therefore to cause a type of "preactivation" of the conditions of the *attention to preferences* elementary behaviours (in particular, only one of these elementary behaviours will be the one that will receive greater preactivation). This it is a typical example of an internal behaviour whose actions are directed to the internal medium. This mechanism constitutes the underlying base of the exploratory behaviour oriented to a specific objective, which is another of the distinguishing properties of the BeCA When there is a total satiety of the internal necessity that caused the activation of the *intero/extero/drive congruence* elemental behaviour ($O_i^E$ = 0), then the level of activation of this elemental behaviour will decrease to zero ($A_i^C$ = 0), independently of the existence of usable external signals. For activation levels of zero or very near to zero the elementary behaviour will not be activated. In this case, the observed external behaviour always will be adaptive.

# 5. Dynamic adjustment of the motivation degree in the action selection

The rule for the modification of the parameter $_i$ is given by expressions (2), (3) and (4). The learning model that controls the dynamic adjustment of the motivation degree (parameter $_i$ in expression 1) can be explained in the following terms:

When for a strong internal state, the external signal able to satisfy this need is not present, then the parameter $_i$ is reinforced in expression 1, so that the action selection begins to be more motivated. Therefore, in later situations to this adjustment, the action selection will begin to give more importance to the internal states. In this case, the exploratory behaviour can be activated for not so strong values of the internal need, although other external signals have been perceived. When an external signal is perceived and the value of the associated internal state to this signal is irrelevant, then the parameter $_i$ of expression 1 is decremented, so that the action selection begins to be lees motivated. Therefore, in later situations to this adjustment, the action selection will begin to give more importance to the available external signals, although the associated internal state to these is not strong. In this case, the exploratory behaviour will need stronger values of the internal need in order to be activated. In any other case the parameter $_i$ is not modified. This cases are represented in expression (2):

$$\alpha_i(t+1) = \begin{cases} f^{+}(\alpha_i(t)) & \text{if } O_i^E > \theta \text{ and } \sum_j Fa_{ij}^S O_j^S = 0 \\ f^{-}(\alpha_i(t)) & \text{if } O_i^E \leq \theta \text{ and } \sum_j Fa_{ij}^S O_j^S > 0 \\ \alpha_i(t) & \text{in other case} \end{cases} \quad (2)$$

where $O_i^E$ is the value of the internal signal from the internal perceptions blackboard level, $\sum_j Fa_{ij}^S O_j^S$ represents the value of all the external signals that are associated to internal state i, and $\theta$ is a threshold value.

The increment for the parameter $_i$ is determined by expression (3). This increment can be seen as a hyperbolic divergence from min, as shown in Figure 3. In expression (3), $\delta$ determines the length of the divergence (how much time it will take to $\alpha_i$ to go from $\alpha$min to $\alpha$max), and $\rho$ determines the speed of the divergence. This smooth modification behaviour simulates an historic memory of the environment (remembered scenario), so that the value of $_i$ is increased only after several iterations in a certain environment.

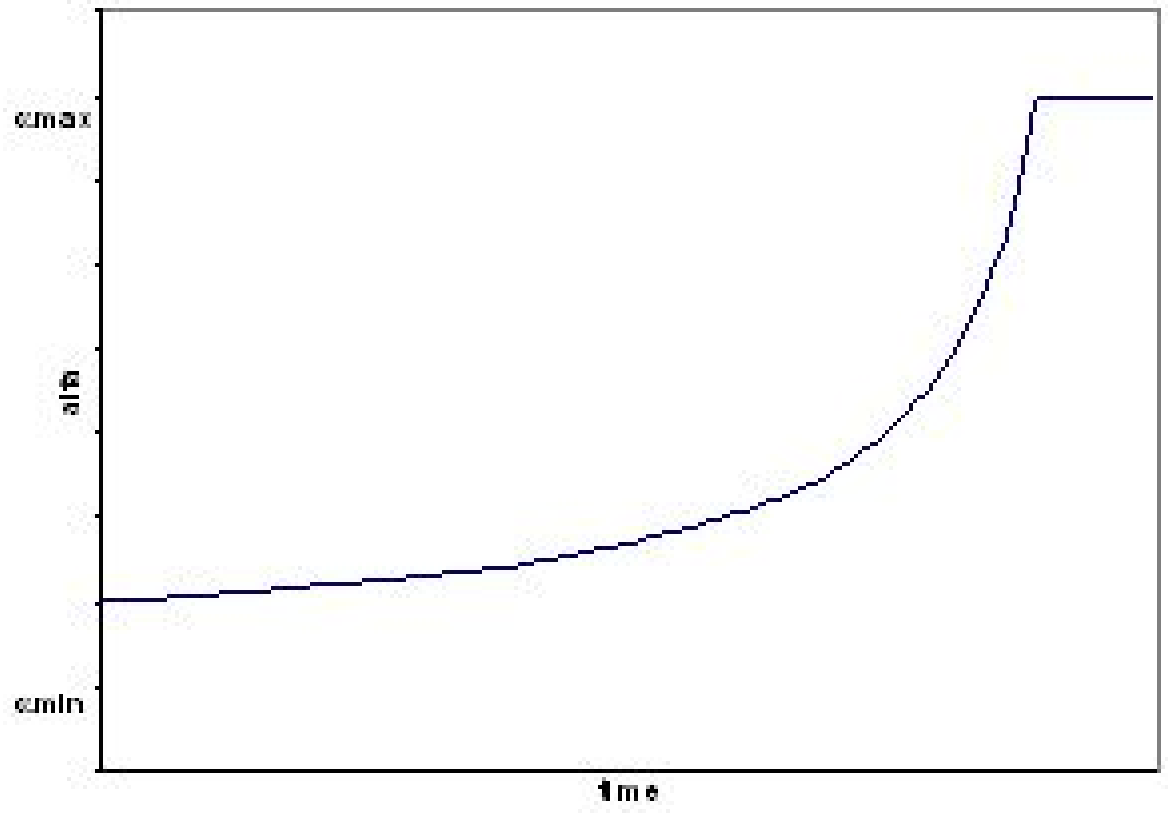

Figure 3. Hyperbolical divergent increment given by expression 3

$$f^{+}(\alpha) = \begin{cases} \left[ -\dfrac{1}{x1(\alpha) - \delta} + \alpha\min - \dfrac{1}{\delta} \right] & \text{if } [\;] < \alpha\max \\ \alpha\max & \text{in other case} \end{cases} \quad (3)$$

$$x1(\alpha) = -\frac{1}{\alpha - \alpha\min + \dfrac{1}{\delta}} + \delta + \rho$$

The decrement of the parameter $_i$ is determined by expression (4). This is similar to the increment described by expression (3), only that it hyperbolically diverges from $\alpha$max, as can be seen in Figure 4.

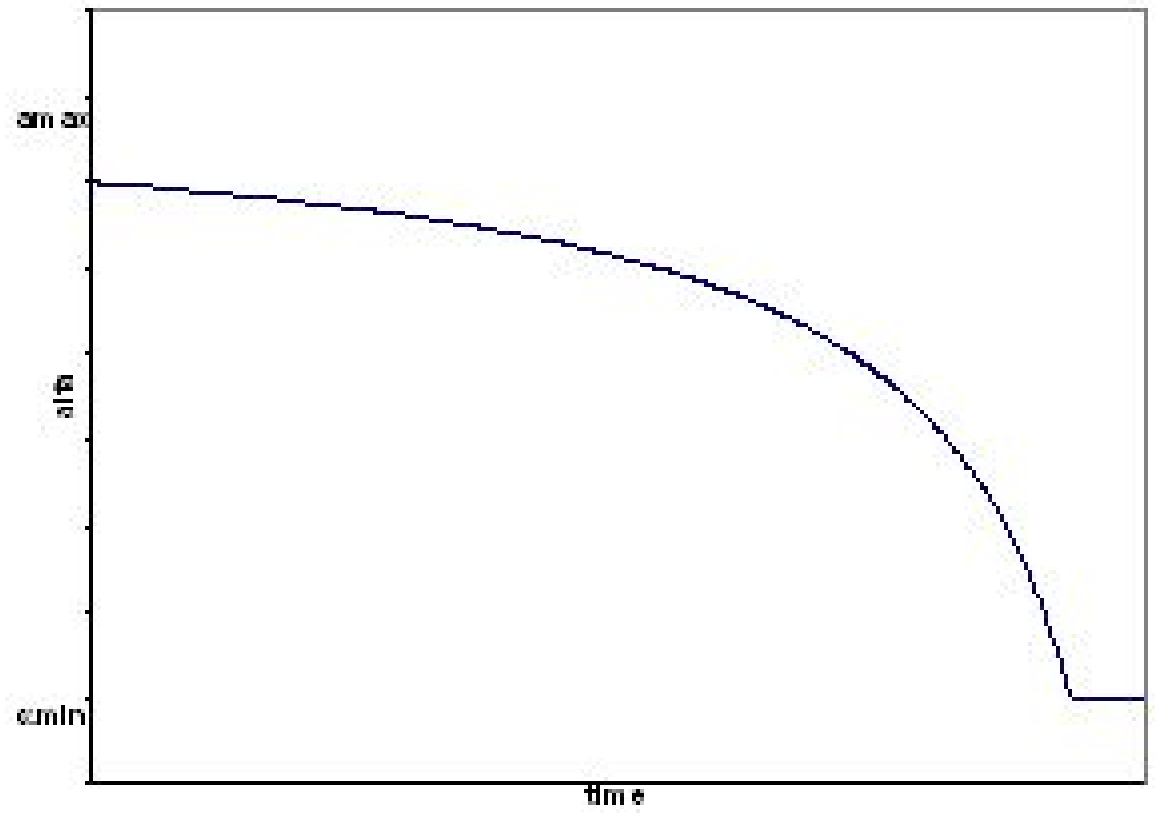

Figure 4. Hyperbolical divergent decrement given by expression 4.

$$f^{-}(\alpha) = \begin{cases} \left[ \dfrac{1}{x2(\alpha) - \delta} + \alpha\max + \dfrac{1}{\delta} \right] & \text{if } [\;] > \alpha\min \\ \alpha\min & \text{in other case} \end{cases} \quad (4)$$

$$x2(\alpha) = \frac{1}{\alpha - \alpha\max - \dfrac{1}{\delta}} + \delta + \rho$$

In Figure 5, we can appreciate an example of the behaviour of the parameter $_i$, as it is increased, decreased, or remains constant, in dependence of the perceived scenario and the internal needs. Note that the increment is faster than the decrement, because the values of $_i$ are closer to $\alpha$max than to $\alpha$min. The graphics of this section were taken from the simulation described in the next section.

The type of learning described in this section can be seen as a particular type of reinforcement learning. This is, the reward or the punishment will be applied to the motivation degree of the action selection, in dependency of how abundant or limited are the signals required in the environment, and how strong or weak are the internal states.

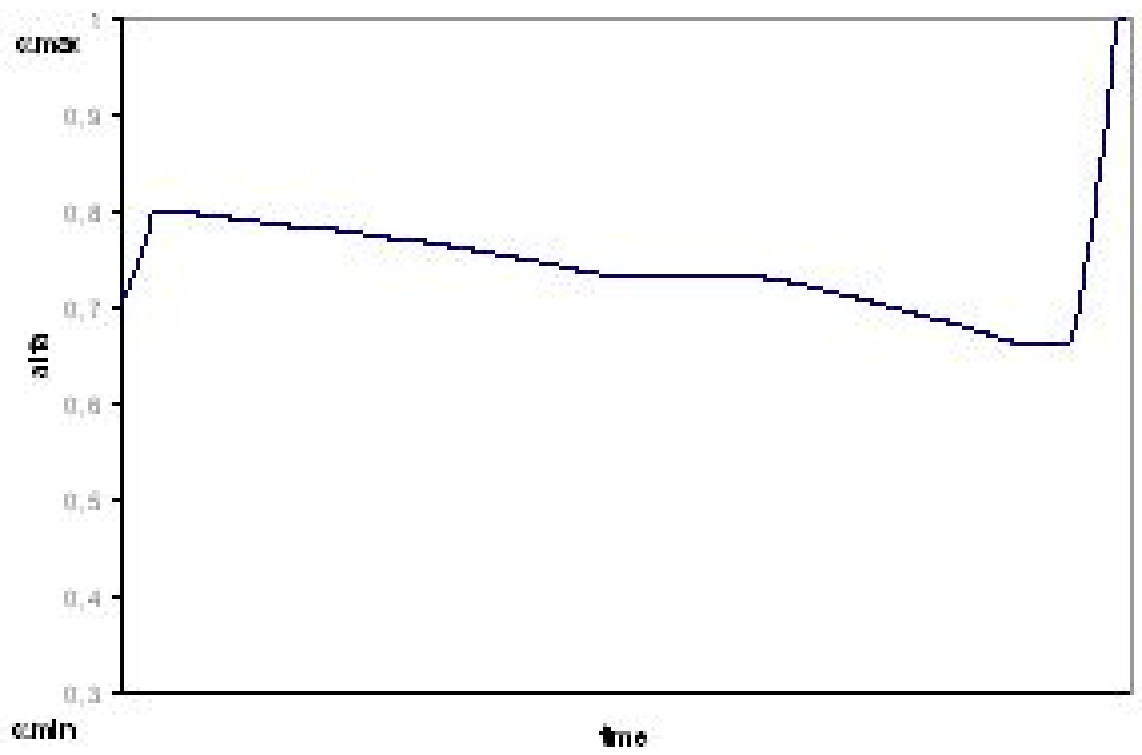

Figure 5. Example of the behaviour of the parameter $_i$

# 6. Simulation and Experiments

## 6.1. The simulation

In order to verify the role of the reinforcement learning in the action selection, a programme was written implementing the behavioural columns architecture, being this used by animats (artificial animals) simulated in virtual reality. The animats were inspired by an original idea developed by Negrete (Negrete and Martinez, 1996).

The graphical representation of the animat is a cone with a semicircumscript sphere. He has primitive motor and perceptual systems developed specially for this simulation. The animat has internal states to satisfy, such as hunger, thirst and fatigue, and the qualities of strength and lucidity; which are affected by the diverse behaviours that he can execute. The implemented behaviours are: wander, explore, approach a specific signal, avoid obstacles, rest, eat, drink, and runaway from a specific signal.

An environment was also simulated, in which the animat develops. In this environment, obstacles, sources of food and water, grass, blobs (aversive stimuli), and spots of different magnitudes can be created, which animat can use to satisfy his internal necessities. This simulation can be accessed via Internet in the address http://132.248.11.4/~carlos/asia/bvl

The simulation allows to modify the different parameters of the animat, for example, to regulate the speed of conditioning or reinforcement.

## 6.2. The experiments

Several experiments were developed to test the role of the adaptation by learning in the action selection (Gershenson et. al., 2000). The experiments discussed here were the ones that we considered most relevant to show how the dynamic varying of the motivation degree (a type of reinforcement learning) improves the action selection in an autonomous agent.

We considered an initial state as shown in Figure 6. We had two animats in separated environments: one with abundant sources of food, and the other one with none. Other stimuli were created in both environments. Both animats were initialized with the same parameters: very high hunger, and no other internal need, with an initial value of $\alpha$ for the hunger column of 0.7.

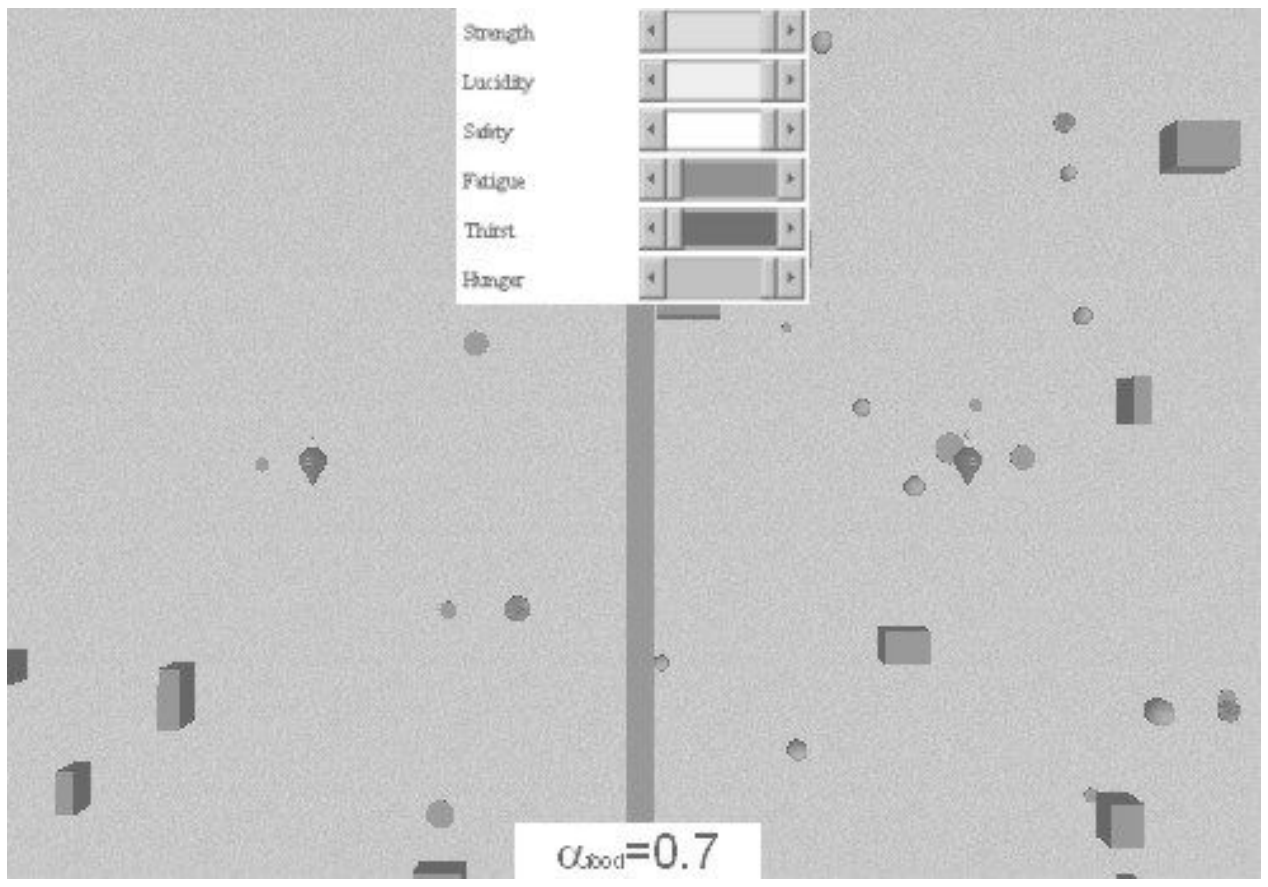

Figure 6. Initial state of the experiment.

Figure 7 shows what occurred after some time of execution of the simulation. The animat in the scarce environment explored in search of food, but found none. This remembered scenario lead to the incremental adjustment of his respective $\alpha_i$ until it reached $\alpha$max. On the other hand, the abundant environment allowed the animat in it to satisfy his hunger quickly. Once his hunger was satiated, the animat wandered, while his respective $\alpha_i$ decreased.

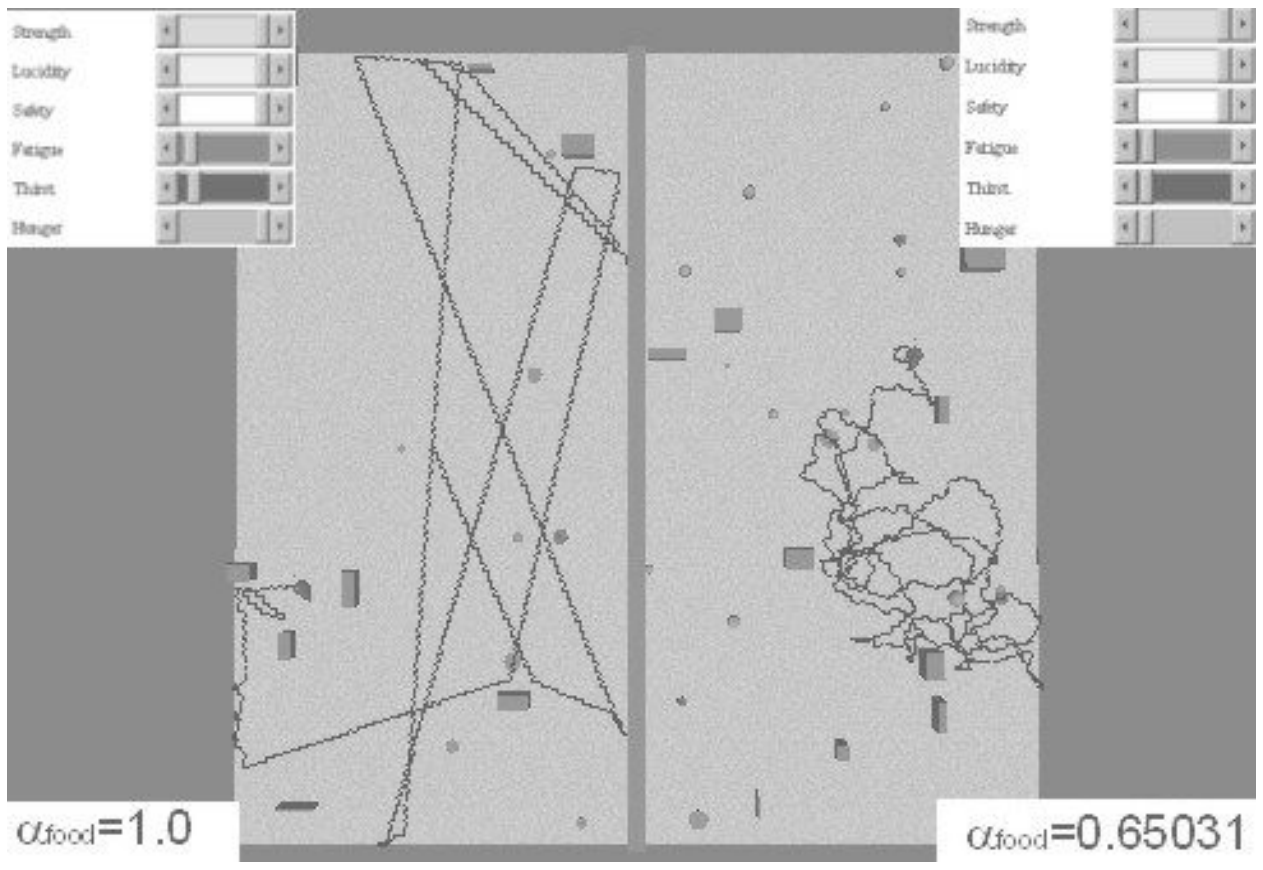

Figure 7. Behaviour patterns described by the animats in their environments.

In this way, we obtained one animat with an $\alpha_i$ equal to $\alpha$max, and another one with an $\alpha_i$ equal to $\alpha$min. Then, we set them with the same degree of hunger, and the animat with an $\alpha_i$ equal to $\alpha$max, since he learned that food is scarce and will be difficult to find, will begin to explore. Meanwhile, the animat with an $\alpha_i$ equal to $\alpha$min will wander, because he learned that he will find abundant food with low effort, and thus allows himself to wander, and will not explore until his hunger is higher. These behaviours can be seen in Figure 8.

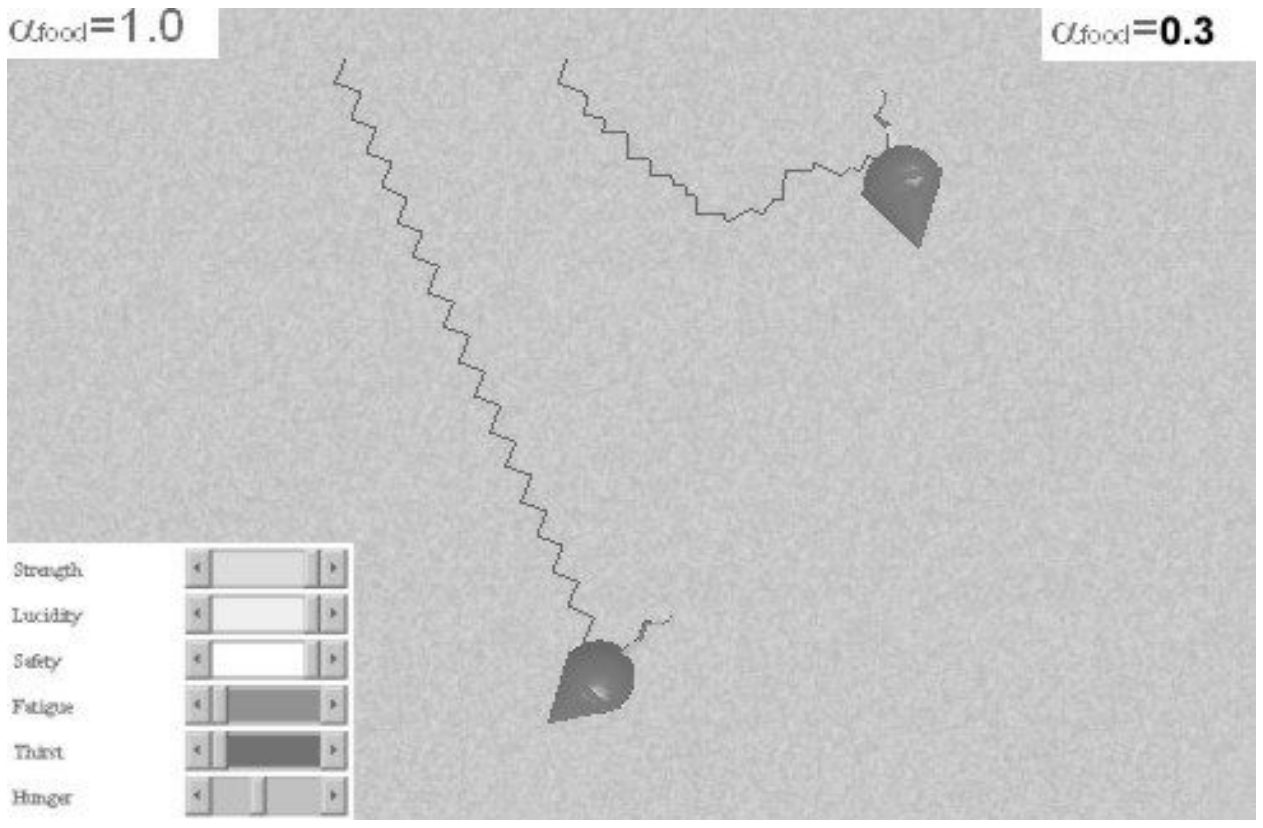

Figure 8. Behaviour patterns of animats with different values of $\alpha_i$ for hunger

With these experiments we could corroborate that with the dynamic adjustment of the motivation degree in the combination of external and internal signals, an autonomous agent will learn certain qualities of its actual dynamic environment, making its behaviour more adaptive.

## 7. Conclusions

In this paper we have discussed and shown how the action selection in an autonomous agent has been improved with adaptation by learning. The type of learning discussed here, reinforcement learning, was used to dynamically adjust the motivation degree in a model for combination of external and internal stimuli incorporated in an ASM previously developed by us.

The dynamic variation of parameter $_i$ in the combination model (expression 1) does the action selection more adaptive, allowing the autonomous agent to develop in an environment from which he possesses certain knowledge. It is important to note that the representation of this knowledge lies in one parameter, not in symbolic representation of knowledge.

## 8. References

Brooks, R. A. (1986) A robust layered control system for a mobile robot. *IEEE Journal of Robotics and Automation*. RA-2, April, pp. 14-23.

Brooks, R. A. (1989) A robot that walks: Emergent behaviour from a carefully evolved network. Neural Computation, 1, 253-262.

Gershenson, C., González, P.P. and Negrete, J. (2000). Action Selection Properties in a Software Simulated Agent. *Proceedings of the 1st Mexican International Conference on Artificial Intelligence (MICAI'2000)*, México.

González, P.P. and J. Negrete (1997) REDSIEX: A cooperative network of expert systems with blackboard architectures. *Expert Systems*, 14(4), 180-189.

González, P.P. (2000). *Redes de Conductas Internas como Nodos Pizarrón: Selección de Acciones y Aprendizaje en un Robot*. Ph.D. Dissertation. Instituto de Investigaciones Biomédicas, UNAM, México.

González, P.P., Negrete, J., Barreiro, A. and Gershenson, C. (2000). A Model for Combination of external and Internal Stimuli in the Action Selection of an Autonomous Agent. *Proceedings of the 1st Mexican International Conference on Artificial Intelligence (MICAI'2000)*, México.

Humphrys, M (1996) Action Selection Methods using Reinforcement Learning. In P. Maes et. al. (Eds), *From Animals to Animats: Proceedings of the Fourth International Conference on Simulation of Adaptive Behaviour* MIT Press/Bradford Books.

Maes, P. (1990) A bottom-up mechanism for behaviour selection in an artificial creature. In J.A. Meyer and S.W. Wilson (Eds), *From Animals to Animats: Proceedings of the First International Conference on Simulation of Adaptive Behaviour* MIT Press/Bradford Books.

Maes, P. (1994) Modeling Adaptive Autonomous Agents. *Journal of Artificial Life*, 1(1,2), MIT Press.

McFarland, D. (1990). What is means for robot behaviour to be adaptive. In J.A. Meyer and S.W. Wilson (Eds), *From Animals to Animats: Proceedings of the First International Conference on Simulation of Adaptive Behaviour* MIT Press/Bradford Books.

Meyer, J.A. and Guillot, A. (1990). Simulation of adaptive behaviour in animats: Review and Prospect. In J.A. Meyer and S.W. Wilson (Eds), *From Animals to Animats: Proceedings of the First International Conference on Simulation of Adaptive Behaviour* MIT Press/Bradford Books.

Negrete, J. and M. Martínez (1996) Robotic Simulation in Ethology. *Proceedings of the IASTED International Conference: Robotics and Manufacturing,* Honolulu, Hawaii, USA, pp. 271-274.

Negrete, J. and P.P. González (1998) Net of multi-agent expert systems with emergent control. *Expert Systems with Applications,* 14(1) 109-116.

Tyrrell, T. (1993) *Computational Mechanisms for Action Selection*. PhD. Dissertation. University of Edimburgh.